\pdfoutput=1

\documentclass[11pt]{article}

\usepackage{acl}

\usepackage{microtype}
\usepackage{times}
\usepackage{latexsym}
\usepackage{soul}
\usepackage{amsmath}
\usepackage{booktabs}
\usepackage{bm}
\usepackage{cleveref}
\usepackage{graphicx}
\usepackage{tabularx}
\usepackage{soul}
\usepackage{xcolor}
\usepackage{todonotes}
\usepackage{multirow}
\usepackage{xpatch}
\usepackage{blindtext}
\usepackage{fdsymbol}
\usepackage{microtype}
\usepackage{hyperref}
\usepackage{adjustbox}
\usepackage{textcomp}
\usepackage{xparse}
\usepackage{enumitem}
\usepackage{makecell}
\usepackage{subcaption}
\usepackage{colortbl}

\usepackage{xparse}
\usepackage{stfloats}
\usepackage{highlight}
\usepackage[linesnumbered,vlined,ruled]{algorithm2e}

\crefformat{section}{\S#2#1#3} %
\crefformat{subsection}{\S#2#1#3}
\crefformat{subsubsection}{\S#2#1#3}

\newcolumntype{P}[1]{>{\centering\arraybackslash}p{#1}}

\newcommand{\samsum}[1]{\textsc{SAMSum}}
\newcommand{\dialogsum}[1]{\textsc{DialogSum}}
\newcommand{\mixandmatch}[1]{\textsc{MixAndMatch}}
\newcommand{\confit}[1]{\textsc{ConFiT}}
\newcommand{\ctrldiasumm}[1]{\textsc{CtrlDiaSumm}}
\newcommand{\cods}[1]{\textsc{CODS}}
\newcommand{\modelshort}[1]{\textsc{C2TFec}}
\newcommand{\modelshortda}[1]{\textsc{ZeroFEC-DA}}
\newcommand{\datashort}[1]{\textsc{Chocolate}}
\newcommand{\pgn}[1]{\textsc{PGN}}
\newcommand{\cliff}[1]{\textsc{CLIFF}}
\newcommand{\conseq}[1]{\textsc{ConSeq}}
\newcommand{\vescore}[1]{\textsc{ChartVE}}

\definecolor{c2}{RGB}{218,0,0}

\definecolor{lightblue}{RGB}{212, 235, 255}
\definecolor{lightorange}{RGB}{255, 204, 168}
\definecolor{lightyellow}{RGB}{255, 255, 168}
\definecolor{lightred}{RGB}{255, 168, 168}
\definecolor{darkred}{RGB}{234, 107, 102}
\definecolor{darkerblue}{RGB}{103, 136, 184}
\definecolor{lightgreen}{RGB}{144, 238, 144}

\definecolor{gold}{rgb}{0.83, 0.69, 0.22}
\sethlcolor{lightblue}
\newcolumntype{Y}{>{\centering\arraybackslash}X}

\iftrue

\NewDocumentCommand{\steeve}
{ mO{} }{\textcolor{gold}{\textsuperscript{\textit{Steeve}}\textsf{\textbf{\small[#1]}}}}
\NewDocumentCommand{\heng}
{ mO{} }{\textcolor{red}{\textsuperscript{\textit{Heng}}\textsf{\textbf{\small[#1]}}}}
\NewDocumentCommand{\ken}
{ mO{} }{\textcolor{purple}{\textsuperscript{\textit{Ken}}\textsf{\textbf{\small[#1]}}}}

\NewDocumentCommand{\mingyang}
{ mO{} }{\textcolor{blue}{\textsuperscript{\textit{Mingyang}}\textsf{\textbf{\small[#1]}}}}
\NewDocumentCommand{\yi}
{ mO{} }{\textcolor{lightgreen}{\textsuperscript{\textit{Yi}}\textsf{\textbf{\small[#1]}}}}
\NewDocumentCommand{\zhenhailong}
{ mO{} }{\textcolor{magenta}{\textsuperscript{\textit{Zhenhailong}}\textsf{\textbf{\small[#1]}}}}
\NewDocumentCommand{\lingyu}
{ mO{} }{\textcolor{violet}{\textsuperscript{\textit{Lingyu}}\textsf{\textbf{\small[#1]}}}}
\NewDocumentCommand{\chang}
{ mO{} }{\textcolor{orange}{\textsuperscript{\textit{Chang}}\textsf{\textbf{\small[#1]}}}}

\else
\newcommand{\steeve}[1]{}
\newcommand{\heng}[1]{}
\newcommand{\mingyang}[1]{}
\newcommand{\yi}[1]{}
\newcommand{\zhenhailong}[1]{}
\newcommand{\lingyu}[1]{}
\newcommand{\chang}[1]{}
\fi
\definecolor{gold}{rgb}{0.83, 0.69, 0.22}

\title{Do LVLMs Understand Charts? \\Analyzing and Correcting Factual Errors in Chart Captioning} %

\author{Kung-Hsiang Huang$^{1, 2*}$ ~~~Mingyang Zhou$^{3}$ ~~~ Hou Pong Chan$^{4*}$\\
{\bfseries ~~~ Yi R. Fung$^{1}$ ~~~ Zhenhailong Wang$^{1}$   ~~~ Lingyu Zhang$^{3}$ ~~~ Shih-Fu Chang$^{3}$ ~~~ Heng Ji$^{1}$}\\
$^{1}$University of Illinois Urbana-Champaign ~~~ $^{2}$Salesforce AI Research \\
$^{3}$Columbia University ~~~ $^{4}$DAMO Academy, Alibaba Group  \\
$^{1}$\texttt{\{yifung2, wangz3, hengji\}@illinois.edu} ~ $^{2}$\texttt{kh.huang@salesforce.com} \\
$^{3}$\texttt{\{mz2974, lz2814, sc250\}@columbia.edu}  ~ $^{4}$\texttt{houpong.chan@alibaba-inc.com}\\
}
\begin{document}
\maketitle
{\def\thefootnote{*}\footnotetext{Work was done while Kung-Hsiang was at UIUC and Hou Pong was at the University of Macau.}}
\begin{abstract}
Advances in large vision-language models (LVLMs) have led to significant progress in generating natural language descriptions for visual contents. %
These powerful models are known for producing texts that are factually inconsistent with the visual input. While some efforts mitigate such inconsistencies in natural image captioning, the factuality of generated captions for structured visuals, such as charts, has not received as much scrutiny. %
This work %
introduces a comprehensive typology of factual errors in generated chart captions. 
A large-scale human annotation effort provides insight into the error patterns in captions generated by various models, ultimately forming the foundation of a dataset, \datashort~. Our analysis reveals that even advanced models like GPT-4V frequently produce captions laced with factual inaccuracies. To combat this, we establish the task of Chart Caption Factual Error Correction and introduce \vescore~, a  visual entailment model that outperforms current LVLMs in evaluating caption factuality. Furthermore, we propose \modelshort~, an interpretable two-stage framework that excels at correcting factual errors. %
This work inaugurates a new domain in factual error correction for chart captions, presenting a novel evaluation metric, and demonstrating an effective approach to ensuring the factuality of generated chart captions. The code and data as well as the continuously updated benchmark can be found at: \url{https://khuangaf.github.io/CHOCOLATE/}.%

\end{abstract}

\section{Introduction}

Large vision-language models (LVLMs) have recently shown impressive capabilities in generating natural language descriptions of visual content like images, videos and charts %
\cite{openai2023gpt4v, google2023bard, liu2023improvedllava,actionpatch2023}. %
Chart captioning is particularly important for data analysts, business analysts, and journalists who rely on accurate chart interpretations for decision-making and reporting. However, no prior work has studied the \textit{factuality}\footnote{Factuality is also known as the \textit{faithfulness} or \textit{factual consistency} between inputs and outputs} of the generated captions. %
Given that factuality is vital for credibility in applications of chart captioning in news articles \cite{liu-etal-2021-visual}, educational resources \cite{fu2022doc2ppt}, and social media \cite{monteiro2017situational}, examining the truthfulness of generated captions is a critical concern.

\begin{figure*}[t]
    \centering
    \includegraphics[width=0.95\linewidth]{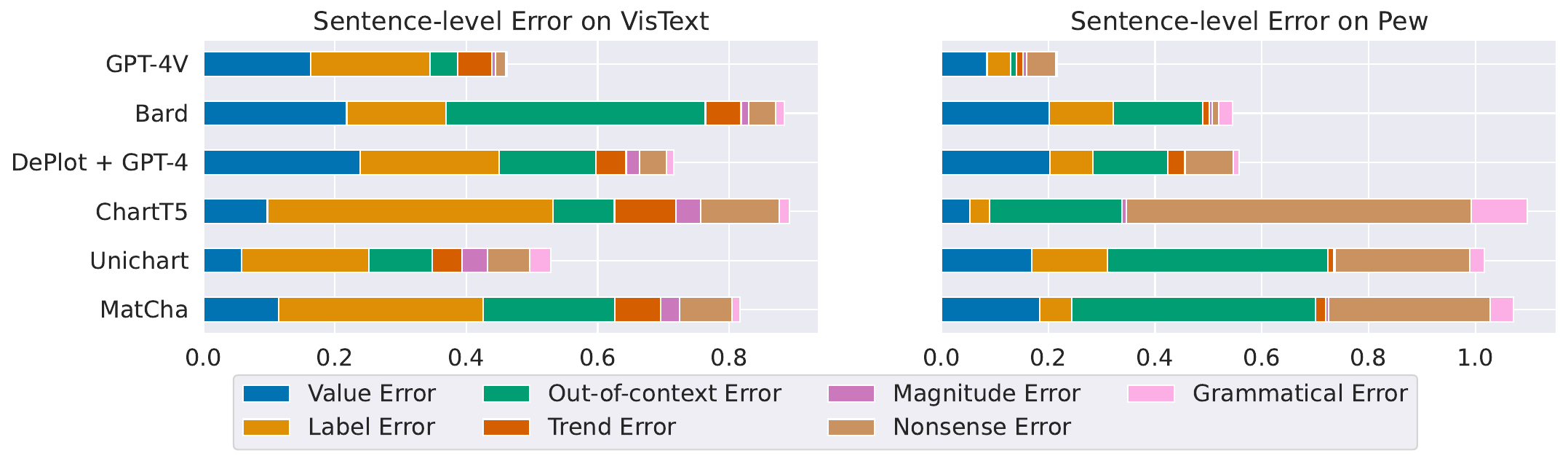}
    \caption{Error distribution for different models on VisText and Pew. The error rates are computed per sentence. An error rate of 0.4 indicates that 40\% of the sentences in the generated captions contain such an error. Note that a single caption may contain multiple types of errors; hence, the maximum value for a stacked bar is greater than 1.0. We show that even the most advanced LVLM, GPT-4V, generates captions with a high rate of factual error.\looseness=-1}%
    \label{fig:error_distribution}
\end{figure*}

To understand the factual errors in chart captioning models, we introduce a typology of factual errors for the chart domain. %
Using this scheme, we conduct a large-scale human annotation study to analyze the distributions of various error types, such as Value Error and Label Error, in captions from various models, from task-specific fine-tuned models to LVLMs (see \Cref{tab::error_typology}). %
The annotated samples are then categorized into three splits, \textsc{Lvlm} (Large-vision Language Models), \textsc{Llm} (Large Language Models), and \textsc{Ft} (Fine-tuned Vision-language Models),  %
based on the architecture and the scale of the underlying models, and form a dataset which we named \datashort~. %
With this dataset collected, we aim to answer three main research questions. First, \textbf{are state-of-the-art chart captioning models able to produce factual captions? We find the answer is no} (\Cref{sec:dataset}). Specifically, 82.06\% of the generated captions are non-factual (see \Cref{tab:dataset_stats}). Even state-of-the-art LVLMs like GPT-4V \cite{openai2023gpt4v} produce a great portion of errors in its generated captions (see \Cref{fig:error_distribution}). %

The prevalence of factual inconsistencies observed in the generated captions by various models underscores the urgent need to mitigate the factual errors of such models. Hence, we introduce a new task, \textit{Chart Caption Factual Error Correction} (\Cref{sec:task}), which presents a novel challenge of rectifying factual inaccuracies in chart captions generated by LVLMs. A pertinent question that arises from this task is: \textbf{how to automatically evaluate the factual consistency between charts and captions?} To tackle this question, we %
present \vescore~, novel visual entailment approach to assess the factual consistency of chart captions. This model is trained by repurposing existing resources from chart summarization and chart question answering. Results show that \vescore~ performs competitively with proprietary LVLMs and outperforms the most advanced open-source LVLM, despite being 64 times less in size.%

Now that we have set up the task, we turn to the challenge of \textbf{how to effectively correct factual errors in chart captions?} We propose \modelshort~ (\Cref{sec:method}), an interpretable two-step framework that decomposes visual reasoning into image-to-structure rendering and text-based reasoning. \modelshort~ first transforms the input chart into a structured data table representation. Grounded in this extracted tabular data, the second component then identifies and fixes any factual inconsistencies in the generated caption through an interpretable reasoning process. Our experiments demonstrate that this explicit decomposition enables more reliable factuality corrections compared to end-to-end approaches. The intermediate symbolic representation acts as an effective bridge between charts and captions, enabling \modelshort~ to significantly outperform competitive baselines including GPT-4V (\Cref{sec:results}). %

In summary, our contributions are as follows: %
\begin{itemize}[noitemsep,nolistsep,leftmargin=*]
    \item We present the first analysis of factual errors in captions produced by models of various scales using a novel error typology, which results in the \datashort~ dataset.
    \item We introduce the Chart Caption Factual Error Correction task that challenges models to correct factual errors in generated chart captions. %
    \item We present \vescore~, a reference-free evaluation metric based on visual entailment that correlates better with human judges than LVLMs. %
    \item We propose \modelshort~, an interpretable two-stage error correction framework that performs better than all existing LVLMs.    
\end{itemize}

\section{Analyzing Factual Errors}
\label{sec:dataset}

To understand the capabilities of existing models in summarizing key information from charts, we conduct a large-scale analysis on six most advanced chart captioning models on the VisText \cite{tang-etal-2023-vistext} and Pew \cite{kantharaj-etal-2022-chart} datasets. To facilitate this process, we introduce an error typology, as illustrated in \Cref{subsec:error_typology}. Upon gathering human annotations, we present a detailed analysis of different captioning models (\Cref{subsec:captioning_model_analysis}) and discuss the quality of the collected data (\Cref{subsec:dataset_quality}). 

\subsection{Error Typology}
\label{subsec:error_typology}
To understand the frequency of various types of errors made by chart captioning systems, we define a typology of errors as detailed below and demonstrate examples in \Cref{tab::error_typology}. %

\begin{table*}[t]
    \small
    \centering
    {
    \begin{tabular}{p{0.27\linewidth} p{0.17\linewidth} p{0.4\linewidth}}
        \toprule

        \textbf{Chart} & \textbf{Category} & \textbf{Example Caption} \\
        \midrule
        \multirow{7}{*}{\includegraphics[width=0.95\linewidth]{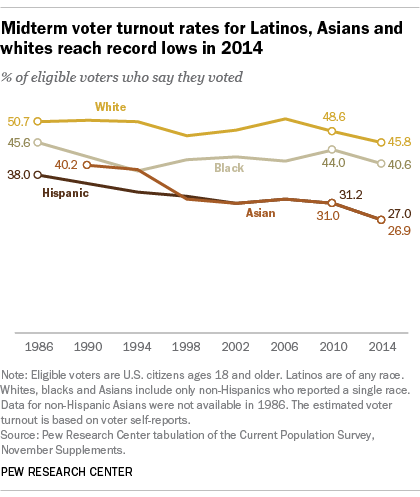}} & Value Error  & Asians have a turnout rate of \textit{\textcolor{red}{20.4\%}} in 1990.\\
        \cmidrule(lr){2-3}
        & Label Error  & \textit{\textcolor{red}{Asians}} have the highest turnout rates across the years.\\
        \cmidrule(lr){2-3}
        &Trend Error  & From 1986-2014, the turnout rates are \textit{\textcolor{red}{increasing}} overall.\\
        \cmidrule(lr){2-3}
        & Magnitude Error  & From 1986-2014, the turnout rates are \textit{\textcolor{red}{sharply}} decreasing overall.\\
        \cmidrule(lr){2-3}
        & Out-of-context Error  & \textit{\textcolor{red}{Vietnamese}} have the highest turnout rates among Asians.\\
        \cmidrule(lr){2-3}
        & Nonsense Error  & From 1986-2014,  \textit{\textcolor{red}{\#?sep \#sep \#sep \#sep}}.\\
        \cmidrule(lr){2-3}
        
        & Grammatical Error  & The turnout rates are \textit{\textcolor{red}{decrease}} overall.\\

        \bottomrule
    \end{tabular}
    }
    \vspace{-2mm}
    \caption{Typology of errors illustrated with an example chart.}%
    \label{tab::error_typology}
    \vspace{-5mm}
\end{table*}

\paragraph{Value Error} A quantitative data value from the chart is incorrectly stated in the caption. This includes numbers representing values on axes, percentages, or other numerical data points.

\vspace{-3mm}
\paragraph{Label Error} A non-numerical label, category, or text element from the chart is incorrectly referenced in the caption. This includes labels on axes, legend items, categorical variables, etc.%

\vspace{-3mm}
\paragraph{Trend Error}  The overall direction of change over time or comparison between groups is incorrectly described in the caption, such as stating an increasing trend when it is actually decreasing.
\vspace{-3mm}
\paragraph{Magnitude Error} The degree or amount of difference described for a trend is unfaithful to the chart, such as stating an increase ``sharp'' when the chart shows it is actually ``smooth''. \looseness=-1%
\vspace{-3mm}
\paragraph{Out-of-context Error}  Concepts, variables, or any information introduced in the caption that does not exist at all in the content of the chart. The caption contains factual statements not grounded in the actual chart contents.
\vspace{-3mm}
\paragraph{Nonsense Error} The caption contains incomplete sentences, disconnected phrases that do not connect logically, or sequences of words that simply do not make coherent sense.
\vspace{-3mm}
\paragraph{Grammatical Error} There are grammatical mistakes in the structure or syntax of the caption.\footnote{Note that we do not consider grammatical errors as factual inconsistency. They are analyzed for assessing fluency.} 

\subsection{Captioning Model Analysis}
\label{subsec:captioning_model_analysis}

We consider various types of models. First, ChartT5 \cite{zhou-etal-2023-enhanced}, MatCha \cite{liu-etal-2023-matcha}, and UniChart \cite{masry2023unichart} are the most advanced task-specific models fine-tuned with in-domain data from the VisText and Pew datasets. Second, DePlot + GPT-4 \cite{liu-etal-2023-deplot, openai2023gpt4} is a LLM-based pipeline approach. Finally, GPT-4V and Bard\footnote{We tested Bard before Gemini's release \cite{google2023gemini}.} are the strongest LVLMs. For each model and dataset, we randomly sample 100 chart figures and generate the corresponding captions. Invalid output sequences, such as empty strings, are filtered out. \looseness=-1

We compute the percentage of sentences with factual errors for different models and datasets, with a breakdown of different error types. Error rates are computed at the sentence level instead of the caption level since different models generate captions of different lengths. A sentence-level evaluation helps mitigate this discrepancy and facilitates a fairer comparison. 

From \Cref{fig:error_distribution}, we made the following observations. First, \textbf{SOTA chart captioning models often fail to produce factual captions}. Additionally, as shown in \Cref{tab:dataset_stats}, we calculated the percentage of non-factual captions, revealing that 82.06\% of captions contain at least one factual error. More importantly, even models like GPT-4V and Bard, which have demonstrated proficiency in a variety of vision-language tasks, produce factually incorrect captions 81.27\% of the time, as recorded in \Cref{tab:dataset_split_stats}. These findings highlight the inherent difficulties of chart captioning tasks and the limitations of SOTA vision-language models.

Second, \textbf{task-specific chart captioning models and LVLMs show opposite trends on the two datasets}. %
Task-specific models, including ChartT5, MatCha, and UniChart, produce fewer errors on the VisText dataset. Conversely, LVLMs, including GPT-4V and Bard, generate significantly fewer errors on the Pew dataset. The key distinctions on these datasets are two: (1) the prevalent labeled values on charts from Pew and (2) the simpler structures in charts from VisText. We hypothesize that LVLMs may be better at utilizing the labeled numbers, while task-specific effectively interpret values via axis alignment. We show an example to validate this hypothesis in \Cref{fig:value_labeling_impact}. %

Third, \textbf{LVLMs cannot consistently outperform task-specific fine-tuned models}. Despite their extensive training data and parameters, LVLMs may be surpassed by task-specific models with appropriate pre-training objectives and architectures. For example, on the VisText dataset, UniChart outperforms Bard and is comparable to GPT-4V in terms of producing more factual captions owing to UniChart's various pre-training objectives for chart comprehension, enabling better interpretation of the relationship between data points within charts. %

The dataset resulting from the analysis is named \datashort~ (\textbf{C}aptions \textbf{H}ave \textbf{O}ften \textbf{C}h\textbf{O}sen \textbf{L}ies \textbf{A}bout \textbf{T}he \textbf{E}vidence), where each instance consists of a chart, a generated chart caption, and error types labeled by human annotators. %
Drawing insights from \citet{tang-etal-2023-understanding} that factual errors produced by different kinds of models may be easier or more difficult to identify, we categorize \datashort~ into three splits: the \textbf{\textsc{Lvlm}} split, with captions from GPT-4V and Bard; the \textbf{\textsc{Llm}} split, featuring DePlot + GPT-4 outputs; and the \textbf{\textsc{Ft}} split, for ChartT5, UniChart, and MatCha captions. Split details are in \Cref{apx:dataset_details}. %

\subsection{Dataset Quality}
\label{subsec:dataset_quality}
To evaluate the quality of \datashort~, we measured inter-annotator agreement by calculating Fleiss' Kappa $\kappa$ \cite{fleiss1971measuring} and the majority vote agreement percentage $p$, in line with the metrics used by \citet{pagnoni-etal-2021-understanding}. %
We applied these metrics across all 5,323 sentences in \datashort~. For determining factual consistency between chart sentences and their corresponding charts, we achieved a Fleiss' Kappa of $\kappa = 0.63$ and a majority vote agreement of $p = 91\%$. For context, \citet{pagnoni-etal-2021-understanding} reported a Fleiss' Kappa of $\kappa = 0.58$ and a majority agreement level of $p = 91\%$. This suggests that \datashort~ exhibits a quality on par with well-established benchmarks in text-based factual inconsistency detection.

\begin{table}[t]
    \small
    \centering
    \begin{adjustbox}{max width=0.47\textwidth}
    {
    \begin{tabular}{lccc}
        \toprule
        
        & \textbf{\# Factual} & \textbf{\# Non-factual} & \textbf{\# Total}\\
        \midrule
        Sentence & 2,561 & 2,762 & 5,323\\
        Caption & 213 & 974 & 1,187 \\

        \bottomrule
    \end{tabular}
    }
    \end{adjustbox}
    \vspace{-2mm}
    \caption{Statistics of the captions we analyzed. A sentence is considered factual if and only if it does not contain any factual error. A caption is considered factual if all its sentences are factual.\looseness=-1} 
    \vspace{-5mm}
    \label{tab:dataset_stats}
    
\end{table}
\section{The Chart Caption Factual Error Correction Task}
\label{sec:task}
The dataset collected in \Cref{sec:dataset} enables us to study the Chart Caption Factual Error Correction task. In this section, we first formally provide the definition of this task (\Cref{subsec:task_definition}) and propose an effective reference-free evaluation metric based on chart visual entailment (\Cref{subsec:ve_score}).

\subsection{Task Definition}
\label{subsec:task_definition}

The input to our task is a chart $\mathcal{E}$ and chart caption $\mathcal{C}$ that may or may not be factually consistent with $\mathcal{E}$. The goal of chart caption factual error correction is to produce a corrected caption $\hat{\mathcal{C}}$ that fixes factual errors in $\mathcal{C}$ with the minimum amount of edits. If $\mathcal{C}$ is already faithful to $\mathcal{E}$, models should output the original caption (i.e. $\hat{\mathcal{C}}$ = $\mathcal{C}$). Following prior work on text-based factual error correction \cite{thorne-vlachos-2021-evidence, huang-etal-2023-zero, gao-etal-2023-reference}, corrections should be made with as few substitution, insertion, and deletion operations as possible since one can trivially achieve $0\%$ non-factual rate by deleting all words in a caption.

\subsection{Reference-free Evaluation With Chart Visual Entailment}
\label{subsec:ve_score}
There was no established metric for evaluating the factual consistency between a chart and the corresponding chart caption. In addition, since our dataset does not contain annotated reference captions\footnote{Reference captions are not collected due to the challenges of curating high-quality references through crowd-sourcing.}, text-based metrics cannot be adopted. %
As a solution, we propose \vescore~, a reference-free evaluation metric based on chart visual entailment, as detailed in the following paragraphs.

\paragraph{\vescore~ Overview}
We formulate the inconsistency detection problem as a chart visual entailment task. Given a chart caption sentence $c$ and a chart $\mathcal{E}$, the task is to predict whether the relationship from $\mathcal{E}$ to $c$ as \textsc{Entailment} (factually consistent) or \textsc{NotEntailment} (factually inconsistent). The main challenge of learning a visual entailment model for this task is the lack of data. To overcome this challenge, we repurpose data from relevant tasks, such as chart QA, as positive samples. Then, we propose a table-guided negative data generation to produce negative samples.%

\paragraph{Positive Data Creation}
We consider datasets from two tasks that are closely related to the chart visual entailment task: chart question answering and chart captioning. %
We utilize two datasets from chart question answering: ChartQA \cite{masry-etal-2022-chartqa} and PlotQA \cite{Methani_2020_PlotQA}. Using a QA2Claim model \cite{huang-etal-2023-zero}, we transform the question-answer pairs into declarative statements and pair them with the original charts to form positive instances (\textsc{Entailment}).
For chart captioning, captions from VisText \cite{tang-etal-2023-vistext} and Chart-to-Text \cite{kantharaj-etal-2022-chart} are segmented into individual sentences. Each sentence is paired with the relevant chart to create a positive instance. These methods allow us to repurpose existing resources for training \vescore~.%

\paragraph{Table-guided Negative Data Generation} %
Generating negative training samples is achieved by perturbing the positive instances grounded in the underlying data tables of the charts. For a chart $\mathcal{E}_i$ and its underlying data table $\mathcal{A}_{{\mathcal{E}_i}}$, we locate values in $\mathcal{A}_{{\mathcal{E}_i}}$ that matches a substring within the positive caption $c^{+}_i$. When a match is found, the substring in the caption is substituted with a different value from the same column in $\mathcal{A}_{{\mathcal{E}_i}}$, yielding a value or label-error infused negative sentence $c^{-}_i$, maintaining relevance while ensuring inconsistency with $\mathcal{E}_i$. %
For trend-related errors, we replace trend-terms found in $c^{+}_i$ with their opposites, drawing on a specific lexicon of terms like ``increase'' and ``decrease,'' thereby creating trend-contradictory statements.
Furthermore, out-of-context errors are crafted by pairing $\mathcal{E}_i$ with a mismatched caption $c^{+}_j$ from another chart, where $i \neq j$. This simulates captions filled with unrelated data.\looseness=-1%

The above process is illustrated in Algorithm \ref{alg:negative_data_gen}. We use the training, development, and test sets of the repurposed datasets for training, validating, and testing \vescore~. This is vital for ensuring that \vescore~ is free from data contamination in downstream applications. In total, we collected over 595K instances partitioned into training, development, and test splits with a ratio of 522:36:37, respectively. \looseness=-1

\paragraph{Learning \vescore~}

We selected UniChart as our base model, given its superior performance amongst comparable-size models\footnote{Our fine-tuning begins with this checkpoint: \url{https://huggingface.co/ahmed-masry/unichart-base-960}.}. Recognizing that UniChart has been pre-trained on chart question answering tasks, we employ a tailored input template $t$ as follows:
\begin{quote}
    \textit{Does the image entail this statement: ``SENTENCE''?}
\end{quote}
In this template, \textit{SENTENCE} replaces the chart caption sentence $c$. Taking in a chart $\mathcal{E}$ and template $t$ as input, UniChart is fine-tuned to produce the token ``yes'' if the chart $\mathcal{E}$ entails the caption sentence $c$, and ``no'' otherwise using maximum likelihood estimate. %
During inference time, we use the same input format and probe the logits corresponding to the ``yes'' ($l_{\text{yes}}$) and ``no'' ($l_{\text{no}}$) decoder tokens. Following this, we apply the softmax function to convert these logits into an entailment score $s(\mathcal{E}, c)$ that ranges from 0 to 1:

{
\small
\begin{equation}
s(\mathcal{E}, c) = \frac{e^{l_{\text{yes}}}}{e^{l_{\text{yes}}} + e^{l_{\text{no}}}}.
\end{equation}
}
Here, $e$ is the base of the natural logarithm. Finally, we compute the minimum of the entailment scores for all sentences within a caption, denoted by $S(\mathcal{E}, \mathcal{C})$, where $\mathcal{C}$ represents the set of all caption sentences for chart $\mathcal{E}$:

{
\small
\begin{equation}
S(\mathcal{E}, \mathcal{C}) = \min_{c \in \mathcal{C}} s(\mathcal{E}, c).
\end{equation}
}

\paragraph{Meta-evaluation of Different Evaluation Metrics}

\begin{table}[t]
    \small
    \centering
    \begin{adjustbox}{max width=0.47\textwidth}
    {
    \begin{tabular}{lccc}
        \toprule
                & \multicolumn{3}{c}{\textbf{\datashort~}} \\

                \cmidrule(lr){2-4}
        \textbf{Model} & \textsc{Lvlm} & \textsc{Llm} & \textsc{Ft} \\
        \midrule
        \textsc{SummaC} & -0.011 & 0.023 & 0.036\\ 
        \textsc{QAFactEval} & 0.064 & 0.045 & 0.054 \\ 
        \midrule
        LLaVA-1.5-13B & 0.002 & 0.057 & 0.214 \\
        ChartLlama & 0.010 & 0.057 & 0.141 \\
        ChartAssistant-S & 0.015 & 0.057 & 0.036 \\
        Bard & -0.014 &  0.105 & \textbf{0.291} \\
        GPT-4V  &  0.157  & \textbf{0.205} & 0.215 \\
        DePlot + GPT-4 & 0.129 & 0.117 & 0.109\\
        
        \midrule
         
        $\vescore~$ (Ours) & \textbf{0.178} & 0.091 & 0.215 \\ 
        
        \bottomrule
    \end{tabular}
    }
    \end{adjustbox}
    \vspace{-2mm} 
    \caption{Kendall’s Tau correlation of different approaches on the \datashort~ dataset.} 
    \label{tab:detection_performance}
    \vspace{-5mm}
\end{table}
To evaluate the effectiveness of different methods in assessing the factuality of generated captions on the \datashort~ dataset, we employ Kendall's Tau \cite{kendall1938new} to compute the correlation between these methods and human judgments. Given the absence of prior work on factual inconsistency detection methods for chart captions, we compare our \vescore~ with zero-shot capable methods, including DePlot + GPT-4, Bard, GPT-4V, and the leading open-source LVLMs, LLaVA-1.5-13B \cite{liu2023improvedllava}, ChartLlama \cite{han2023chartllama}, and ChartAssistant-S \cite{meng2024chartassisstant}. Text-based factuality metrics, \textsc{SummaC} \cite{laban-etal-2022-summac} and \textsc{QAFactEval} \cite{fabbri-etal-2022-qafacteval}, which compute the factual consistency between the reference caption and the generated caption, are also included. The prompts for these models are detailed in \Cref{apx:prompts}. \looseness=-1

Meta-evaluation, summarized in \Cref{tab:detection_performance}, shows that, overall, \textbf{metrics exhibit the strongest correlation with human judgment on the \textsc{Ft} split and the weakest on the \textsc{Lvlm} split}. This pattern aligns with expectations: the \textsc{Ft} captions are littered with more obvious mistakes, such as out-of-context and nonsense errors, while errors stemming from LVLMs are harder to detect since they often demand intricate inferences regarding the data points' positions relative to the axes, as detailed in \Cref{fig:error_distribution}. Importantly,  Our \vescore~ excels on the challenging \textsc{Lvlm} split, but less so on the \textsc{Llm} split, likely due to shifts in token distribution, as DePlot + GPT-4 occasionally employs table-centric terminology (e.g., ``columns'' and ``entries'') absent from \vescore~'s training data. Despite this, \textbf{\vescore~ compares favorably to proprietary LVLMs and outperforms open-source LVLMs, despite \vescore~ being \textit{64 times smaller in scale}.} \looseness=-1

Bard and GPT-4V lead on the \textsc{Llm} and \textsc{Ft} splits, respectively. However, Bard shows a negative correlation on the \textsc{Lvlm} split, hinting at LVLMs' limitations in assessing the factuality of chart captions. Thus, we advocate for using the best-performing metric for each split for evaluation.

\section{Methodology}
\label{sec:method}
\begin{figure}[t]
 \centering
 \includegraphics[width=\linewidth]{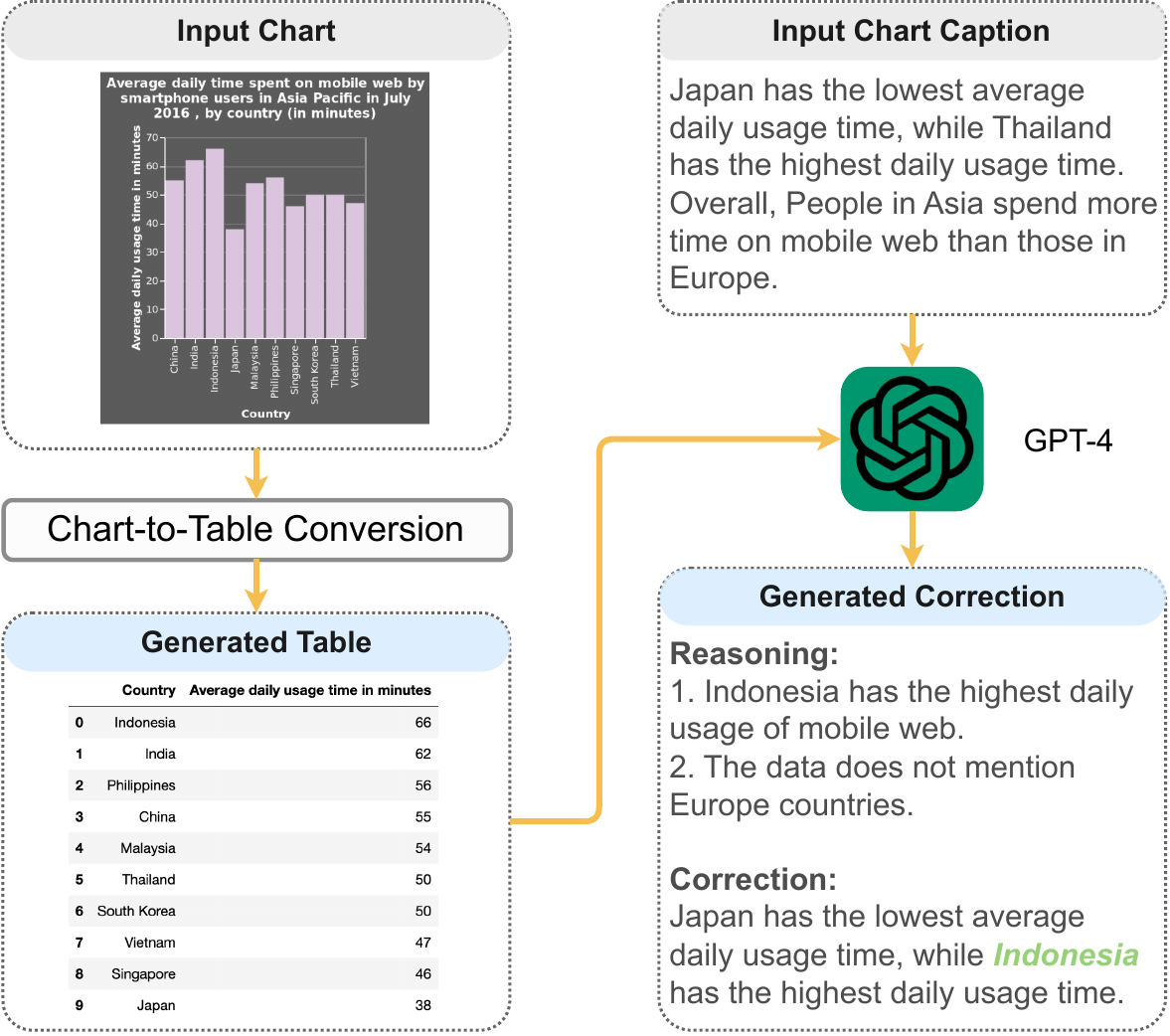}
 \vspace{-2mm}
 \caption{An overview of \modelshort~. Our approach decomposes visual reasoning into image-to-structure rendering and text-based reasoning, allowing for interpretability and better correction of chart captions.\looseness=-1}
 \vspace{-5mm}
 \label{fig:method_overview}
\end{figure}

In correcting factual errors in generated captions, we propose \modelshort~, a two-step, interpretable framework, as shown in \Cref{fig:method_overview}. \modelshort~ first transforms input charts into data tables (\Cref{subsec:chart2table}), then rectifies errors in the caption using the tabular data (\ref{subsec:table_based_correction}). This framework is motivated by our analysis on ``DePlot + GPT-4'', which shows that a notable proportion of errors in caption generation originated from the DePlot component. To mitigate this, we develop a stronger chart-to-table model based on UniChart, significantly improved with expansive fine-tuning datasets. The advantage of \modelshort~ is its ability to harness the reasoning strengths of GPT-4 to faithfully correct errors, boosting caption factuality.\footnote{Here, we do not consider approaches based on LVLMs due to their tendency towards factual errors.}

\begin{table*}[t]
    \small
    \centering
    \begin{adjustbox}{max width=0.98\textwidth}
    {
    \begin{tabular}{lcccccc}
        \toprule
        
        \textbf{Dataset Split $\rightarrow~$}        & \multicolumn{2}{c}{\textbf{\datashort~-\textsc{Lvlm}}} & \multicolumn{2}{c}{\textbf{\datashort~-\textsc{Llm}}} & \multicolumn{2}{c}{\textbf{\datashort~-\textsc{Ft}}} \\

                \cmidrule(lr){2-3}
                \cmidrule(lr){4-5}
                \cmidrule(lr){6-7}
        \textbf{Evaluation Metric $\rightarrow~$} & \vescore~ (\%) & Levenshtein & GPT-4V (\%) & Levenshtein & Bard (\%)  & Levenshtein  \\
        \textbf{Correction Model $\downarrow~$}\\
        \midrule
        N/A                   & 31.13 &  0.0  & 23.47 &  0.0  &     43.10   & 0.0   \\
        LLaVA                 & 31.20 & 19.09 & 22.45 & ~9.20 & 52.94 & 16.94 \\
        Bard                  & 14.13 &127.83 & 31.77 & 77.63 & 75.69 & 42.80 \\
        GPT-4V                & \underline{33.30} & 31.26 & \textbf{52.35} & 50.57 & \underline{76.55}  & 30.92\\
        DePlot + GPT-4        & 32.47 & 81.37 & 22.45 & 21.25 & 70.31 & 38.79 \\
        $\text{DePlot}_{\text{CFT}}$ + GPT-4        & 32.91 & 84.99 & 25.51 & 55.35 & 70.47 & 40.12 \\
        \midrule
        \modelshort~ (Ours)                  & \textbf{34.34} & 72.19 & \underline{39.29} & 53.11 & \textbf{81.14} & 37.36\\ 
        
        \bottomrule
    \end{tabular}
    }
    \end{adjustbox}
    \vspace{-2mm}
    \caption{Correction performance of different models on the \datashort~ dataset. \vescore~ measures factuality by computing the entailment probability from each chart to the corresponding caption sentences. GPT-4V and Bard, when used as evaluation metrics, rate each chart caption as factually consistent with the chart or not. Levenshtein computes the edit distance between the corrected caption and the original caption (denoted as ``N/A''). 
    Metric scores are shown separately for each of the three data splits based on captioning model source. The highest and second highest performing models per evaluation metric and split are highlighted in boldface and underlines respectively.} %
    \label{tab:main_results}
    \vspace{-5mm}
\end{table*}

\subsection{Chart-To-Table Conversion} 
\label{subsec:chart2table}
The training data for our chart-to-table model is sourced from datasets including VisText, Chart-to-Text, ChartQA, and PlotQA, where we repurpose original charts and underlying data tables for our model's training. We collected a total of 65K instances with a train:dev:test split of 61:2:2. Similar to DePlot \cite{liu-etal-2023-deplot}, our model is also trained to generate chart titles, enhancing its ability to contextualize the data represented in table form. Let $\mathcal{M}$ denote our proposed model. For a given chart figure $\mathcal{E}$, the model autoregressively generates a chart title $\mathcal{T}$ and a corresponding table $\mathcal{A}$ (i.e. $\mathcal{T}, \mathcal{A} = \mathcal{M}(\mathcal{E})$).

\subsection{Table-based Error Rectification}
\label{subsec:table_based_correction}
With the input chart now converted into structured tabular data, the second phase uses the reasoning capacity of LLMs to address the factual inconsistency between $\mathcal{C}$ and the generated table $\mathcal{A}$. Here, we use GPT-4 as the LLM. GPT-4 first provides an explanatory breakdown of detected factual errors in $\mathcal{C}$ based on the table contents. It then uses this explanation to produce a corrected caption $\mathcal{\hat{C}}$. This transparent process enables users to validate the reasoning behind each correction.

\modelshort~ separates the factual verification from language generation, taking advantage of the complementary strengths of separate vision and language models tailored to their respective domains. The symbolic table representation acts as a bridge to enhance and validate factual consistency in chart captions.

\section{Experimental Settings}

To assess \modelshort's ability in factual error correction for chart captions, we experiment on the \datashort~ dataset.

\paragraph{Datasets} Our \datashort~ dataset includes 1,187 chart-caption pairs with factually consistent and inconsistent captions, as detailed in \Cref{sec:dataset}. It is split into \textsc{Lvlm}, \textsc{Llm}, and \textsc{Ft}, reflecting the diversity of models that generated the captions.

\paragraph{Baselines} Since \datashort~ does not comprise training data, we compare \modelshort~ against zero-shot capable LVLMs and LLMs, including LVLMs, LLaVA-1.5-13B, GPT-4V, Bard, as well as DePlot + GPT-4. For a fairer comparison between our approach and DePlot, we continue fine-tuning DePlot for an additional 5,000 steps on VisText, an approach which has been shown effective for adapting models to unseen domains \cite{huang-etal-2023-zero}. We denote this model as $\text{DePlot}_{\text{CFT}}$. The prompts used for each model are described in \Cref{apx:prompts}. %

\paragraph{Evaluation Metrics} We assess the factual consistency between corrected captions and input charts using \vescore~, GPT-4V, and Bard, according to our recommendations in \Cref{subsec:ve_score}. In addition, since corrections should be made with as few edits as possible, we measure the number of edits using the Levenshtein distance \cite{levenshtein1966binary}. %

\section{Results}
\label{sec:results}

\subsection{Main Results}
The results in \Cref{tab:main_results} demonstrate that our \modelshort~ achieves the best performance for factual consistency on the \textsc{Lvlm} and \textsc{Ft} splits, and takes the second place on the \textsc{Llm} split. This indicates that \textbf{the two-step process of first transforming charts into structured data tables and then rectifying factual inconsistencies using table-caption alignment is an effective strategy.} %

Additionally, we see that \modelshort~ outperforms the pipeline approaches of DePlot/$\text{DePlot}_{\text{CFT}}$ + GPT-4 across the board. While both methods utilize an intermediate tabular representation and leverage GPT-4 for language generation/correction, \modelshort~ employs a superior chart-to-table conversion model with much more comprehensive training datasets. This results in extracted tables that more faithfully capture the underlying chart data, better facilitating the downstream factual error correction. \modelshort~ also requires a relatively small number of edits to captions according to Levenshtein distance, making focused changes to improve factuality while minimizing revisions. An example output from \modelshort~ is shown in \Cref{fig:qualitative_analysis}. By comparison, the proprietary LVLM Bard produces corrected captions requiring 127.83 as many character-level edits on average. This signals excessive rewriting rather than targeted error correction. After manually inspecting Bard's outputs, we found the reason is that Bard oftentimes try to improve the fluency of the caption by paraphrasing. Hence, it makes more edits to the generated captions.

\begin{figure*}[t]
    \centering
    \includegraphics[width=0.95\linewidth]{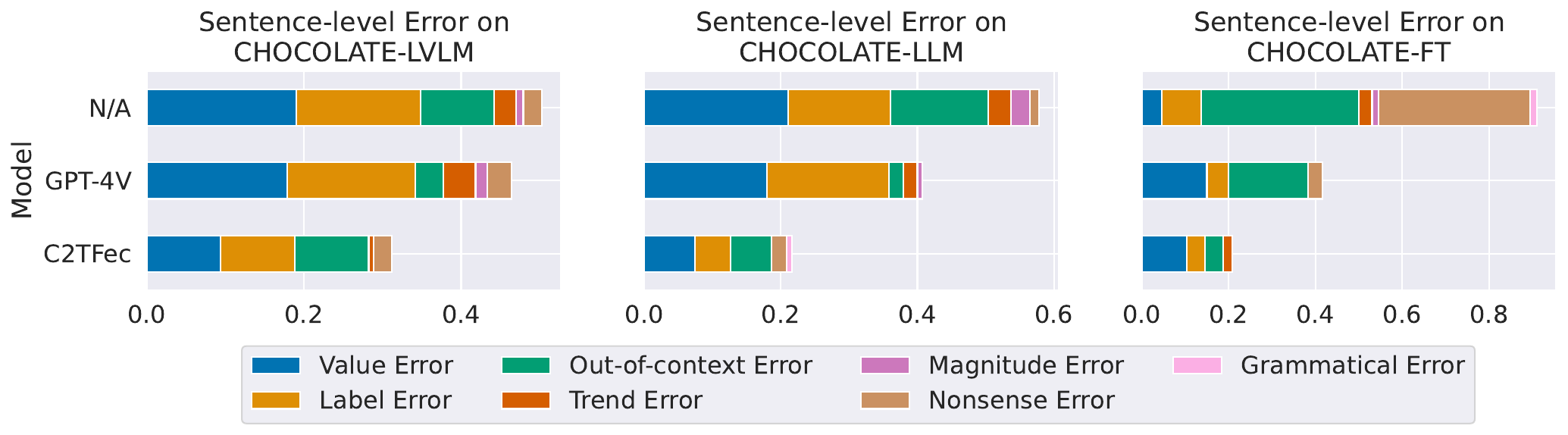}
    \vspace{-2mm}
    \caption{Human evaluation results on subsets of the \datashort~ dataset, comparing \modelshort~ and GPT-4V. \modelshort~ corrects significantly more errors compared to GPT-4V, especially Value, Label, and Trend Errors.} 
    \vspace{-5mm}
    \label{fig:human_eval}
\end{figure*}

Bard's underperformance on the \textsc{Lvlm} split and its negative correlation with human judgments of factuality, as shown in \Cref{tab:detection_performance}, implies its unreliability in detecting errors in chart captions. Additionally, when used as an evaluator, GPT-4V tends to assign %
high factuality scores to its own corrected outputs on all three splits (see \Cref{tab:gpt_4v_eval}), while other metrics show GPT-4V lagging behind \modelshort~. This suggests GPT-4V may suffer from the \textit{self-enhancement bias} \cite{zheng2023judging}, overestimating its own performance when used for evaluation. We thus perform human evaluations in \Cref{subsec:human_eval} to verify the effectiveness of our approach.

\subsection{Human Evaluation}
\label{subsec:human_eval}
Our human assessments focus on comparing \modelshort~ with GPT-4V by using the same annotation tasks detailed in \Cref{sec:dataset} for factual error identification, with the same annotators evaluating. We sampled 30 charts from each split of \textsc{Lvlm}, \textsc{Llm}, and \textsc{Ft}. For each chart, human judges are presented with a caption generated by one of the models.

\begin{figure*}[t]
    \centering
    \includegraphics[width=0.95\linewidth]{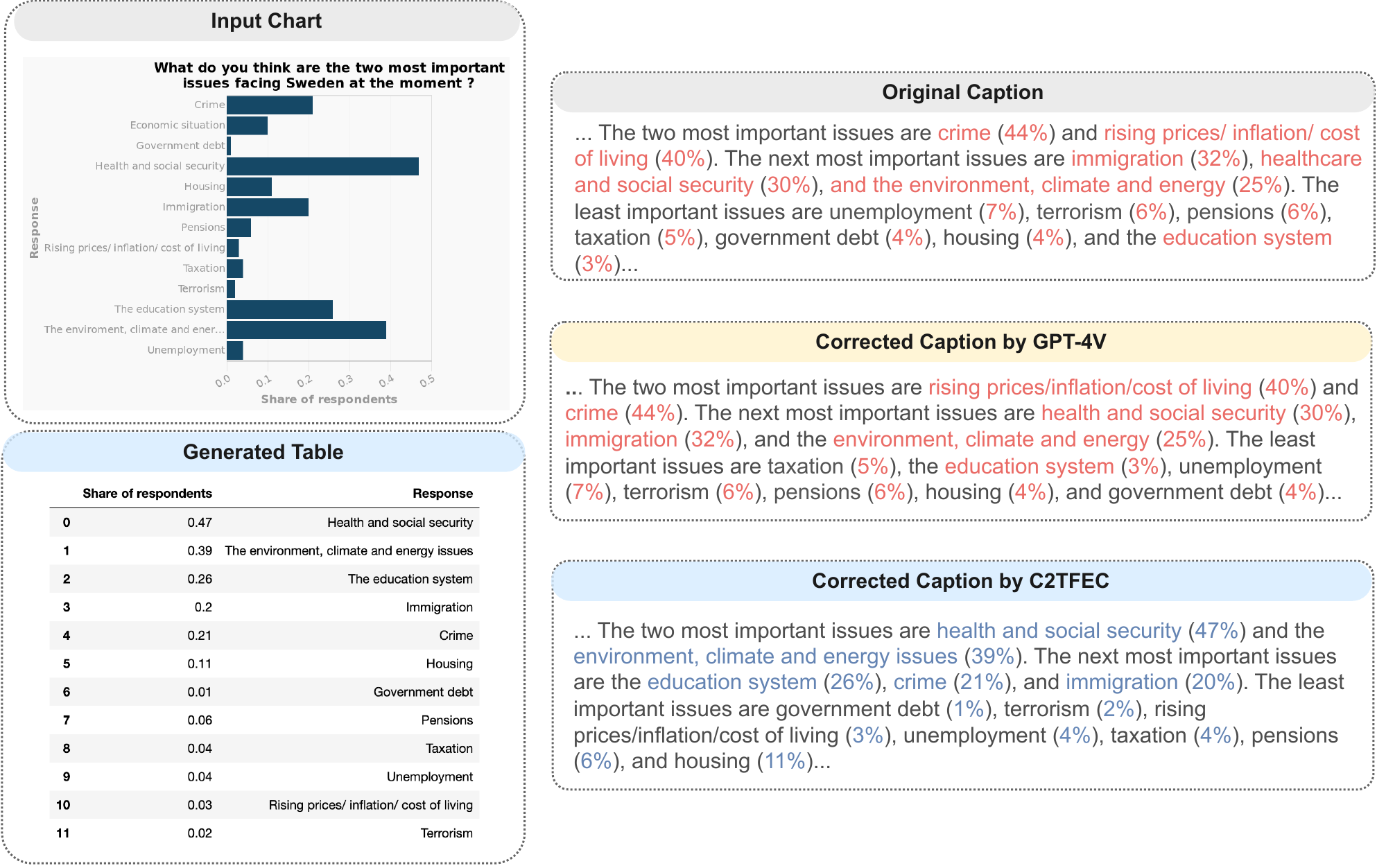}
    \vspace{-2mm}
    \caption{An example showing how decomposing the visual reasoning process into image-to-structure rendering and text-based reasoning allows \modelshort~ to accurately rectify errors in chart captions. Texts marked in \textcolor{darkred}{red} indicate non-factual information units in the caption, whereas those marked in \textcolor{darkerblue}{blue} represent information units faithful to the chart. In this instance, \modelshort~ successfully corrects all Value and Label Errors presented in the original caption. Conversely, GPT-4V fails to identify the factual inconsistencies and merely reorders the entities in the caption. } 
    \vspace{-5mm}
    \label{fig:qualitative_analysis}
\end{figure*}

\Cref{fig:human_eval} demonstrates \modelshort's superiority in multiple error categories, especially with a substantial decrease in Value Errors, over 20\% better in the \textsc{Lvlm} and \textsc{Llm} splits, and halving the overall error rate compared to GPT-4V. \modelshort~ virtually eliminated Trend Errors, highlighting its strong error correction ability, particularly for axes-related errors like Label, Value, and Trend errors. A representative comparison is shown in \Cref{fig:qualitative_analysis}. GPT-4V's shortcomings seem to stem from its failure to accurately infer data point values from charts as evidenced in \Cref{fig:gpt4v_extracted_table}.

In contrast, GPT-4V is better in addressing Out-of-context Errors, involving information out of the chart's scope. However, GPT-4V seemed challenged in rectifying errors within captions generated by itself, particularly within the \textsc{Lvlm} split. This observation echoes recent findings on LLMs' inability to self-correct \cite{huang2023large, valmeekam2023can}, we find that \textbf{LVLMs also cannot perform self-correction}. More importantly, our human evaluation results, combined with our findings in \Cref{tab:main_results} and \Cref{tab:gpt_4v_eval}, reflect that GPT-4V is subject to serious self-enhancement bias.  Consequently, \textbf{although GPT-4V's capabilities are formidable, we recommend not using them to assess their own outputs}. \looseness=-1

\section{Related Work}

\subsection{Chart Captioning}
Chart captioning is essential for accurately interpreting and communicating the information conveyed by chart images, particularly in news articles and social media, where factuality is imperative to prevent misinformation. While current datasets like FigureQA \cite{Kahou2017FigureQAAA}, DVQA \cite{kafle2018dvqa}, PlotQA \cite{Methani_2020_PlotQA}, VisText \cite{tang-etal-2023-vistext}, and Chart-to-Text \cite{kantharaj-etal-2022-chart} offer chart image descriptions and question-answer pairs to train models, advancements in vision-language models like ChartT5 \cite{zhou-etal-2023-enhanced}, MatCha \cite{liu-etal-2023-matcha}, and UniChart \cite{masry2023unichart} have largely prioritized relevance and fluency over factual accuracy. Our work provides a rigorous characterization of factual errors in chart captioning and comparisons of methods to address this gap. By focusing on faithfulness and correction, we complement the emphasis of prior work and aim to produce more trustworthy chart captions.

\subsection{Factual Error Correction}
Prior research in factual error correction has mainly targeted text summarization and fact-checking. Within summarization, the bulk of work %
has been carried out in the news domain and often involves methods that substitute inconsistent entities from the source text. Some studies have enhanced this approach through entity-replacement reranking techniques \cite{chen-etal-2021-improving}, autoregressive models for rewriting and perturbation filtering \cite{cao-etal-2020-factual, zhu-etal-2021-enhancing, adams-etal-2022-learning}, and editing strategies that focus on selective deletion \cite{wan-bansal-2022-factpegasus}. In contrast, \citet{fabbri-etal-2022-improving} employed sentence compression datasets to train their models. More recently, \citet{gao-etal-2023-reference} have expanded the focus of these studies to include dialogue summarization.

Moving to the domain of fact-checking, this area has experienced a flurry of activity, particularly with the increased attention on combating misinformation \cite{fung-etal-2021-infosurgeon, wu-etal-2022-cross, fung2022the, huang-etal-2023-faking, huang2023manitweet, qiu2023amrfact,Huang2024FromPT,Qiu2024VALOREVALHC}. Early approaches train a distantly supervised model that involves a masker and a corrector \cite{shah2020automatic, thorne-vlachos-2021-evidence}. \citet{thorne-vlachos-2021-evidence} made significant strides by developing the first factual error correction dataset for fact-checking, thus enabling fully supervised training for error correctors. Recently, \citet{huang-etal-2023-zero} propose an interpretable framework that breaks down the process of factual error correction into individual components. Our study builds on these insights and extends them to a multimodal context, which challenges models to understand the chart images and the consistency between different modalities. \looseness=-1

\section{Conclusion}

Our study exposes the prevalent issue of factual errors in chart captions generated by various chart captioning models and introduces \datashort~ to scrutinize these errors. We establish the Chart Caption Factual Error Correction task to propel the creation of trustworthy captioning systems and present \vescore~, an evaluation model surpassing LVLMs in mirroring human assessments of caption factuality. Our two-stage correction framework, \modelshort~, provides an interpretable means of improving caption factuality by transforming visual data into structured tables for more faithful error corrections. Our work marks an essential step in ensuring verifiable and trustworthy chart captions. Future directions include extending our approach to multimodal contexts beyond charts, developing more sophisticated error detection and correction algorithms, and creating datasets covering a broader range of visual content.

\section{Ethical Considerations}
Text generation models pre-trained on information from the Web are known to demonstrate various biases. Despite the primary focus on models and datasets that represent the English-speaking population's culture, manual examinations of the \datashort~ dataset reveal no evidence of biases related to gender, age, race, or other socioeconomic factors \cite{qiu-etal-2023-gender,Wang2024NewJN}.

In \Cref{sec:dataset} and \Cref{subsec:human_eval}, we recruited annotators to assess the factual consistency of chart captions. The annotators were fairly compensated for their efforts, as detailed in Appendix \ref{apx:anntation_details}. During the annotation process, we made provisions for open communication, allowing the annotators the flexibility to work at their preferred pace and the freedom to withdraw from the project at any point. Additionally, we took measures to protect the anonymity of the contributors by excluding any personally identifiable information from the dataset.
\section{Limitations}
We acknowledge that our study did not rigorously examine the sensitivity of different systems to the variations in the prompts used. The effectiveness of several natural language processing tasks is known to be influenced by the design of the input prompts. Our omission of a systematic sensitivity analysis means that there could be a range of responses to different prompts that we have not accounted for, which may affect the generalization of our results. However, we did not perform prompt tuning to craft prompts that benefit our proposed model. Therefore, the comparisons across all models are fair. Due to the scope of our study, we leave the prompt sensitivity experiments for future work.

In addition, charts in the datasets we used are mostly line plots and bar plots. Future efforts can extend our work with additional analyses for other types of charts, such as violin plots and distribution plots. %

Moreover, our investigation centered on the factuality of machine-generated chart captions, particularly those produced by LVLMs. We chose not to examine captions written by humans, as our primary objective was to highlight concerns regarding the reliability of automated chart captioning, a tool on which there is an increasing dependence for humans. The analysis of human-generated chart captions represents another avenue for future research, which could offer valuable comparisons and insights into the effectiveness of automated versus human captioning practices.

\section*{Acknowledgement}
This research is based upon work supported by U.S. DARPA SemaFor Program No. HR001120C0123, DARPA ECOLE Program No. HR00112390060, and the Molecule Maker Lab Institute: an AI research institute program supported by NSF under award No. 2019897 and No. 2034562. The views and conclusions contained herein are those of the authors and should not be interpreted as necessarily representing the official policies, either expressed or implied, of DARPA, or the U.S. Government. The U.S. Government is authorized to reproduce and distribute reprints for governmental purposes notwithstanding any copyright annotation therein. 
Hou Pong Chan was supported in part by the Science and Technology Development Fund, Macau SAR (Grant Nos. FDCT/060/2022/AFJ, FDCT/0070/2022/AMJ) and the Multi-year Research Grant from the University of Macau (Grant No. MYRG2020-00054-FST). 
\bibliography{anthology,custom}

\clearpage

\appendix

\section{Further Discussions}

\paragraph{Error Typology} Our error typology was derived from our systematic preliminary analysis of around 50 generated captions. This exploratory phase allowed us to identify common error patterns and categories emergent from the data itself. Moreover, our methodology was enriched by drawing on insights from prior works such as Chart-to-Text, ChartT5, and VisText. 

\paragraph{Model Usage} The GPT-4V version we used was \texttt{gpt-4-vision-preview}. We query GPT-4V via API and the cost for using GPT-4V to produce captions for our \textsc{CHOCOLATE} dataset is less than \$50 USD. For the Bard model, we obtain its outputs during October 2023. There is no cost using Bard since we access it via its web interface. Both models are prompted in single conversation, as seen in \Cref{fig:lvlm_correction_prompt}.

\paragraph{Captioning Model Analysis} In addition the findings we summarize in \Cref{subsec:captioning_model_analysis}, we also found that \textbf{the error distribution for each model differs on different datasets}. Almost all models make significantly more Nonsense Errors on the Pew dataset. In addition, task-specific models observe a non-negligible increase in Out-of-context Errors on the Pew dataset. Both observations could be explained by the fact that these models are sometimes confused about the charts in Pew, which are often associated with more complicated structures.

\begin{table*}[t]
    \small
    \centering
    \begin{adjustbox}{max width=0.95\textwidth}
    {
    \begin{tabular}{lcccccc}
        \toprule
                
        \textbf{Dataset $\rightarrow~$} &  \multicolumn{3}{c}{\textbf{VisText}} &  \multicolumn{3}{c}{\textbf{Pew}} \\

                \cmidrule(lr){2-4}
                \cmidrule(lr){5-7}

    \textbf{Model $\downarrow~$}     & \textbf{\# Factual} & \textbf{\# Non-Factual} & \textbf{Factual Rate (\%)} & \textbf{\# Factual} & \textbf{\# Non-Factual} & \textbf{Factual Rate (\%)} \\ %
    
    \midrule
    ChartT5 & 70 & 197 & 26.22 & 10 & 123 & 7.52 \\
    MatCha & 67 & 107 & 38.51 & 63 & 301 & 17.31 \\
    UniChart & 88 & 67 & 56.77 & 62 & 228 & 21.38 \\
    DePlot & 217 & 223 & 49.32 & 301 & 246 & 55.03 \\
    Bard & 252 & 578 & 30.36 & 438 & 336 & 56.59 \\
    GPT-4V & 361 & 213 & \textbf{62.89} & 632 & 143 & \textbf{81.55} \\

        \bottomrule
    \end{tabular}
    }
    \end{adjustbox}
    \caption{Sentence-level error counts and error rates.} 
    \label{tab:sent_level_error_stats}
\end{table*}
\begin{table*}[t]
    \small
    \centering
    \begin{adjustbox}{max width=0.95\textwidth}
    {
    \begin{tabular}{lcccccc}
        \toprule
                
        \textbf{Dataset $\rightarrow~$} &  \multicolumn{3}{c}{\textbf{VisText}} &  \multicolumn{3}{c}{\textbf{Pew}} \\

                \cmidrule(lr){2-4}
                \cmidrule(lr){5-7}

    \textbf{Model $\downarrow~$}     & \textbf{\# Factual} & \textbf{\# Non-Factual} & \textbf{Factual Rate (\%)} & \textbf{\# Factual} & \textbf{\# Non-Factual} & \textbf{Factual Rate (\%)} \\ %
    
    \midrule
    ChartT5 & 17 & 83 & 17.00 & 7 & 93 & 7.00 \\
MatCha & 33 & 67 & 33.00 & 2 & 98 & 2.00 \\
UniChart & 50 & 46 & \textbf{52.08} & 3 & 97 & 3.00 \\
DePlot & 10 & 86 & 10.42 & 17 & 83 & 17.00 \\
Bard & 1 & 95 & 1.04 & 6 & 93 & 6.06 \\
GPT-4V & 17 & 83 & 17.00 & 50 & 50 & \textbf{50.00} \\

        \bottomrule
    \end{tabular}
    }
    \end{adjustbox}
    \caption{Caption-level error counts and error rates.} 
    \label{tab:sent_level_error_stats}
\end{table*}

Furthermore, in \Cref{fig:error_distribution}, the error rates are computed as the number of such errors divided by the number of sentences. While this provides an overview of the \textit{frequency} for each error, it does not indicate the likelihood of a value/label/trend/magnitude-related mention in the generated captions being factual. This limitation can result in an underrepresentation of certain error types – for instance, the infrequent occurrence of Magnitude Errors as shown in \Cref{fig:error_distribution} is more a consequence of the scarcity of magnitude-related mentions in the captions rather than an indication of the models' superior trend variance comprehension. To address this, we sample 30 generated captions for each model from each dataset and compute another error rate as the number of sentences containing such non-factual mentions over the number of sentences containing such mentions. The results are shown in \Cref{tab:error_rates}. The outcomes corroborate the observations in \Cref{subsec:captioning_model_analysis}, while \Cref{tab:error_rates} offers a supplementary perspective on model performance.

\paragraph{Meta-evaluation Results} For the text-based metrics presented in \Cref{tab:detection_performance}, they both perform weakly in determining the factuality of the generated caption. This is largely because charts often contain much denser information compared to the corresponding reference. As a result, text-only factuality metrics are unsuitable for assessing factual consistency between charts and captions. Additionally, the negative values in \Cref{tab:detection_performance} signal instances where the performance of the methods inversely correlates with human judgments, indicating their poor effectiveness in serving as evaluation metrics for identifying factual errors in generated captions. Furthermore, we experimented with ROC AUC as an additional meta-evaluation metric. As shown in \Cref{tab:detection_performance_auc}, our original conclusion with Kendall's Tau still holds. \looseness=-1

\paragraph{Understanding The Upper Bound} We seek to understand the performance upper bound of our proposed two-stage framework by replacing generated tables with ground-truth data tables. Since the ground-truth data tables in Pew are not available, we experiment with only the instances from the VisText dataset. The results are demonstrated in \Cref{tab:upper_bound}.

\paragraph{Ablation Studies} We compare our chart-to-table component against DePlot, $\text{DePlot}_{\text{CFT}}$, and GPT-4V. Since the Pew split of Chart-to-Text does not contain ground truth labels, we show performance on the VisText test set. We use RMS-F1 proposed in \citet{liu-etal-2023-deplot} as the evaluation metric.  From \Cref{tab:chart2tabl_eval}, we see that our Chart-to-Table component significantly outperforms all baselines, indicating its effectiveness in converting charts into their underlying data tables.

\begin{figure*}[b]
    \centering
    \includegraphics[width=0.9\linewidth]{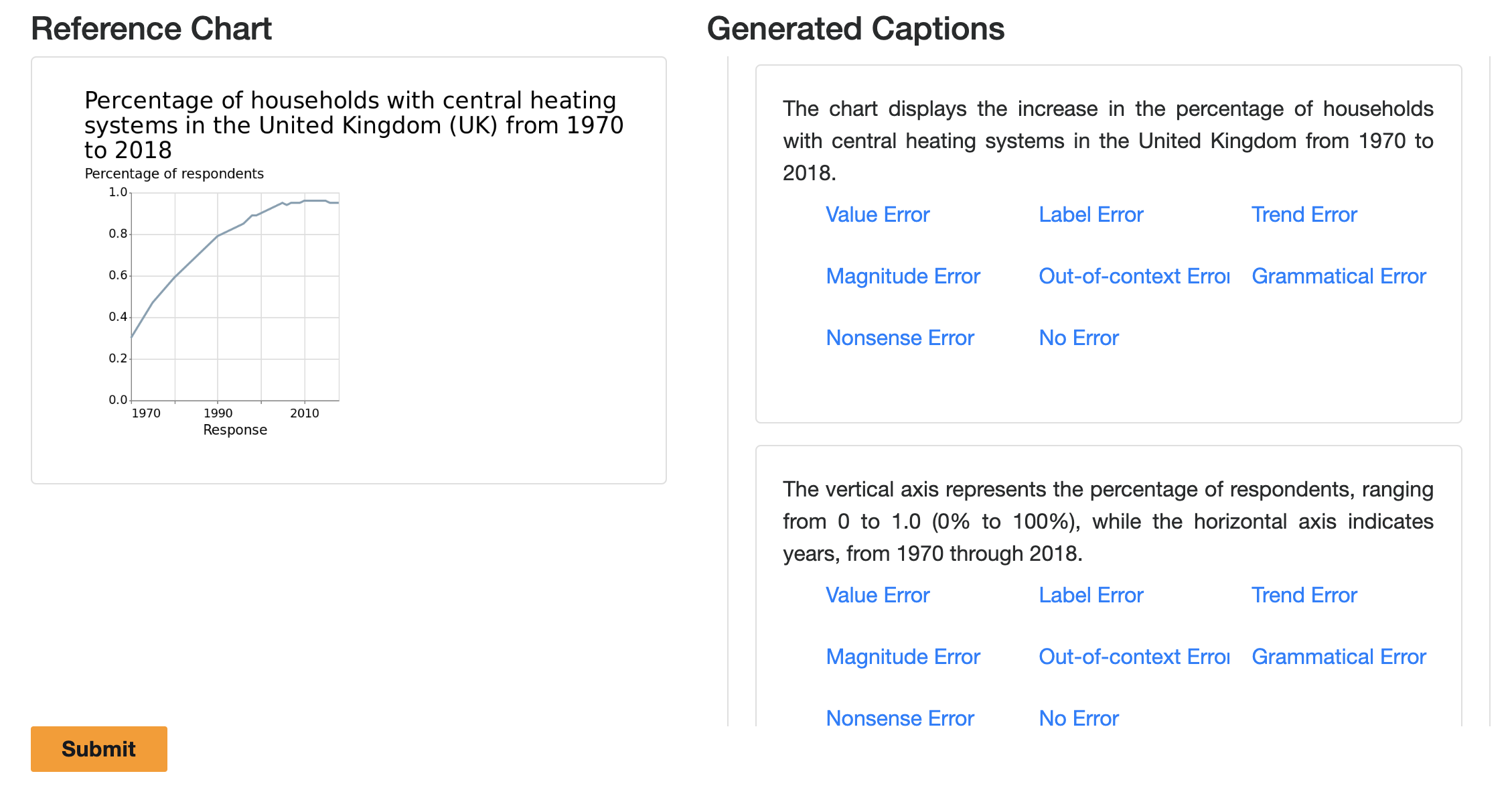}
    
    \caption{Human annotation interface for our data collection discussed in \Cref{sec:dataset}. Examples of each type of error from \Cref{tab::error_typology} are also displayed in the annotation interface. We were not able to show these examples in this figure due to space limits.} 
    
    \label{fig:annotation_interface}
\end{figure*}

\section{Annotation Details}
\label{apx:anntation_details}
In this section, we present the details of our human annotation conducted in \Cref{sec:dataset}.

\subsection{Worker Qualification}

We laid out specific preliminary criteria for the recruitment of MTurk workers with impressive performance records. These prerequisites comprise a HIT approval percentage of 99\% or above, a minimum of 10,000 approved HITs, and the worker's location within the United Kingdom, Canada, or the United States. 

Moreover, beyond these initial criteria, suitable workers have to successfully pass two staged qualification examinations focused on identifying factual errors in generated chart captions. To optimize the qualification procedure, the authors manually annotate two HITs, each consisting of one chart and one caption produced by one of our chart captioning models. In every qualification round, annotators are exposed to one of these annotated examples. Workers whose annotations fail to correspond closely with ours are eliminated from the selection procedure.

Finally, a group of 7 annotators who successfully navigated all three stages of qualification tests were chosen. Additionally, each HIT was meticulously crafted to ensure that annotators could achieve an equivalent hourly pay rate of \$15 - \$20, assuming they work without interruption.

\begin{table*}[t]
    \small
    \centering
    \begin{adjustbox}{max width=0.95\textwidth}
    {
    \begin{tabular}{lcccccccc}
        \toprule
                
        \textbf{Dataset $\rightarrow~$} &  \multicolumn{4}{c}{\textbf{VisText}} &  \multicolumn{4}{c}{\textbf{Pew}} \\

                \cmidrule(lr){2-5}
                \cmidrule(lr){6-9}
         \textbf{Error Type $\rightarrow~$} & Value & Label & Trend & Magnitude & Value & Label & Trend & Magnitude \\
         \textbf{Model $\downarrow~$}\\
        \midrule
        
        ChartT5 & \gca{92.31} (12/13) & \gca{64.71} (33/51) &  \gca{32.00} (8/25) & \gca{100.00} (3/3) 
                & \gca{66.67} (2/3) & \gca{100.00} (2/2) &  N/A (0/0) & N/A (0/0)   \\
        MatCha & \gca{71.43} (5/7) & \gca{50.00} (13/26) &  \gca{23.33} (7/30) &  \gca{50.00} (1/2) 
                & \gca{100.00} (2/2) & \gca{66.67} (2/3) &  N/A (0/0) & N/A (0/0)   \\
        UniChart & \textbf{\gca{33.33} (3/9)} & \textbf{\gca{29.41} (10/34)} &  \textbf{\gca{0.00} (0/14)} &  \gca{50.00} (2/4) 
                & \gca{51.72} (15/29) & \gca{46.67} (14/30) &  \gca{100.00} (1/1)  & N/A (0/0)   \\
        DePlot + GPT-4 & \gca{51.52} (34/66) & \gca{44.78} (30/67) &  \gca{30.77} (8/26) &  \textbf{\gca{0.00} (0/7)}  
                & \gca{49.25} (33/67) & \gca{34.48} (10/29) &  \gca{46.15} (6/13) &  \textbf{\gca{0.00} (0/3)}  \\
        Bard & \gca{69.12} (47/69) & \gca{69.39} (34/49) & \gca{43.75} (14/32) & \gca{15.38} (2/13) 
        & \gca{38.10} (40/105) & \gca{27.71} (23/83) &  \gca{11.11} (2/18)  & \gca{40.00} (2/5)   \\
        GPT-4V  & \gca{40.48} (17/42) &  \gca{33.33} (17/51) & \gca{20.75} (11/53) & \gca{23.53} (4/17) 
                & \textbf{\gca{8.20} (10/122)} & \textbf{\gca{9.02} (11/122)} &  \textbf{\gca{16.67} (2/12)}  & \gca{33.33} (2/6)   \\

        \bottomrule
    \end{tabular}
    }
    \end{adjustbox}
    \caption{Error rates (\%) are calculated by dividing the number of sentences containing such non-factual mentions (e.g. non-factual mentions of values) by the number of sentences containing such mentions (e.g. all mentions of values). The lower the error rate, the better the performance.\looseness=-1} 
    \label{tab:error_rates}
\end{table*}
\begin{table*}[t]
    \small
    \centering
    \begin{adjustbox}{max width=0.98\textwidth}
    {
    \begin{tabular}{lcccccc}
        \toprule
        
        \textbf{Dataset Split $\rightarrow~$}        & \multicolumn{2}{c}{\textbf{\datashort~-\textsc{Lvlm}}} & \multicolumn{2}{c}{\textbf{\datashort~-\textsc{Llm}}} & \multicolumn{2}{c}{\textbf{\datashort~-\textsc{Ft}}} \\

                \cmidrule(lr){2-3}
                \cmidrule(lr){4-5}
                \cmidrule(lr){6-7}
        \textbf{Evaluation Metric $\rightarrow~$} & \vescore~ (\%) & Levenshtein & GPT-4V (\%) & Levenshtein & Bard (\%)  & Levenshtein  \\
        \textbf{Correction Model $\downarrow~$}\\
        \midrule

        \modelshort~                                 & 29.29 & 62.85 & 40.63 & 35.63 &  49.49 & 23.48\\ 
        \modelshort~ (w/ GT Table)                   & \textbf{29.90} & 52.82 & \textbf{40.69} & 32.59 & \textbf{50.93} & 23.47\\ 
        \bottomrule
    \end{tabular}
    }
    \end{adjustbox}
    
    \caption{Correction performance of different models on the \datashort~ dataset. \vescore~ measures factuality by computing the entailment probability from each chart to the corresponding caption sentences. GPT-4V and Bard, when used as evaluation metrics, rate each chart caption as factually consistent with the chart or not. Levenshtein computes the edit distance between the corrected caption and the original caption. Metric scores are shown separately for each of the three data splits. Note that the Bard metric corresponds to Gemini Pro \cite{google2023gemini} since the experiments were conducted after its release. } 
    \label{tab:upper_bound}
\end{table*}

\begin{table}[t]
    \small
    \centering
    \begin{adjustbox}{max width=0.47\textwidth}
    {
    \begin{tabular}{lccc}
        \toprule
                & \multicolumn{3}{c}{\textbf{\datashort~}} \\

                \cmidrule(lr){2-4}
        \textbf{Model} & \textsc{Lvlm} & \textsc{Llm} & \textsc{Ft} \\
        \midrule
        
        LLaVA-1.5-13B &  0.501 & 0.566 & 0.670 \\
        Bard &   0.488 & 0.617 &  \textbf{0.743} \\
        GPT-4V  &  0.638 & \textbf{0.738} & 0.678 \\
        DePlot + GPT-4 & 0.609 & 0.629 & 0.590  \\
        
        \midrule
         
        $\vescore~$ (Ours) & \textbf{0.646} & 0.595 & 0.676 \\ 
        
        \bottomrule
    \end{tabular}
    }
    \end{adjustbox}
    \caption{ROC AUC of different approaches on the \datashort~ dataset.} 
    \label{tab:detection_performance_auc}
\end{table}
\subsection{Annotation Guidelines}

In this task, you will evaluate the factual errors for a generated caption with regard to the reference chart.
To correctly solve this task, follow these steps:
\begin{itemize}[noitemsep,nolistsep]
\item Carefully read the generated caption and the reference chart.
\item Compare the generated caption against the reference chart and decide whether the caption contains any factual error defined below.
\item You should click/press the button if an error occurs. A blue button indicates the caption contains the corresponding factual error, while a white button means the caption does not contain such an error.
\end{itemize}

\textbf{Warning}: Annotations will be checked for quality against control labels, low-quality work will be rejected.

\textbf{Error definition}

\begin{itemize}[noitemsep,nolistsep]
\item \textbf{Value error}: A quantitative data value is incorrect.

\item \textbf{Label error}: A non-quantitative data value is incorrect.

\item \textbf{Trend error}: The direction of a trend is wrong.

\item \textbf{Magnitude error}: The magnitude or variance of a trend is wrong.

\item \textbf{Out-of-context error}: The caption introduces concepts that are not present in the chart.

\item \textbf{Grammatical error}: The grammar of the caption is wrong.

\item \textbf{Nonsense error}: The caption is incomplete or does not make sense at all.
\end{itemize}

\subsection{Annotation Interface}
The interface for our human annotation is shown in \Cref{fig:annotation_interface}.

\section{Dataset Details}
\label{apx:dataset_details}
\Cref{tab:dataset_split_stats} presents the detailed statistics of each split in our dataset.
\begin{table*}[t]
    \small
    \centering
    \begin{adjustbox}{max width=0.95\textwidth}
    {
    \begin{tabular}{lcccccc}
        \toprule

        & \multicolumn{2}{c}{\textbf{\datashort~-\textsc{Lvlm}}} & \multicolumn{2}{c}{\textbf{\datashort~-\textsc{Llm}}} & \multicolumn{2}{c}{\textbf{\datashort~-\textsc{Ft}}}  \\

                \cmidrule(lr){2-3}
                \cmidrule(lr){4-5}
                \cmidrule(lr){6-7}
        & \# Factual & \# Non-factual & \# Factual & \# Non-factual & \# Factual & \# Non-factual\\
        \midrule
        Sentence & 1,683 & 1,270 & 518 & 469 & 360 & 1,023 \\
        Caption & 74 & 321 & 27 & 169 & 112 & 484 \\

        \bottomrule
    \end{tabular}
    }
    \end{adjustbox}
    
    \caption{Dataset statistics per split. A sentence is considered factual if and only if it does not contain any factual error. A caption is considered factual if all its sentences are factual.} 
    \label{tab:dataset_split_stats}
    
\end{table*}

\section{Implementation Details}

\begin{algorithm}[h]

\caption{Table-guided Negative Data Generation}
\label{alg:negative_data_gen}
\KwIn{
    Data table $\mathcal{A}_{\mathcal{E}_i}$ for chart $\mathcal{E}_i$,
    Positive caption sentence $c^{+}_i$.
}
\KwOut{
    Set of negative caption sentences $C^{-}_i = \{c^{-}_{i, \text{value}}, c^{-}_{i, \text{trend}}, c^{-}_{i, \text{context}}\}$.
}

\BlankLine

Initialize $C^{-}_i$ as an empty set\;
Define a lexicon of trend terms $T$\;
Define entailment threshold $\tau$\;
// Generate Value and Label Errors\;
\For{each cell value $v$ in $\mathcal{A}_{\mathcal{E}_i}$}{
    \If{$v$ is a substring of $c^{+}_i$}{
        Randomly sample a new value $v'$ from the same column in $\mathcal{A}_{\mathcal{E}_i}$\;
        Replace $v$ in $c^{+}_i$ with $v'$ to get $c^{-}_{i, \text{value}}$\;
        Add $c^{-}_{i, \text{value}}$ to $C^{-}_i$\;
    }
}
// Generate Trend Errors\;
\For{each trend term $t$ in $T$}{
    \If{$t$ is found in $c^{+}_i$}{
        Replace $t$ in $c^{+}_i$ with its antonym to get $c^{-}_{i, \text{trend}}$\;
        Add $c^{-}_{i, \text{trend}}$ to $C^{-}_i$\;
    }
}
// Generate Out-of-Context Errors\;
Randomly select a different chart $\mathcal{E}_j$ where $j \neq i$\;
Pair $\mathcal{E}_i$ with unrelated caption sentence $c^{+}_j$ to get $c^{-}_{i, \text{context}}$\;
Add $c^{-}_{i, \text{context}}$ to $C^{-}_i$\;

\BlankLine

\Return $C^{-}_i$\;

\end{algorithm}
\begin{table}[t]
    \small
    \centering
    \begin{adjustbox}{max width=0.98\textwidth}
    {
    \begin{tabular}{lc}
        \toprule
        
        \textbf{Model}        & \textbf{RMS-F1}  \\

        \midrule
        DePlot                   & 15.97 \\
        $\text{DePlot}_{\text{CFT}}$ & 78.23 \\
        GPT-4V                  & 10.44 \\
        \midrule
        Chart-to-Table (Ours)   & 83.60 \\ 
        
        \bottomrule
    \end{tabular}
    }
    \end{adjustbox}
    \caption{Chart-to-Table performance comparison on the VisText dataset.} 
    \label{tab:chart2tabl_eval}
\end{table}

\subsection{Details of the Chart-To-Table Model}
Our chart-to-table model takes in as input a graphical chart and outputs a linearized data table format, using \texttt{\textbackslash t} to delimit columns and \texttt{\&\&\&} for row separation. The backbone of our approach is UniChart \cite{masry2023unichart}, due to its diverse chart-oriented pre-training objectives that have demonstrated strong performance on relevant tasks.
\subsection{Table-guided Negative Data Generation}

 In Algorithm \ref{alg:negative_data_gen}, we depict the details of how we generate negative data for our \vescore~ model.
\subsection{Model Training}
The Chart-To-Table model and \vescore~ are optimized using AdamW for a maximum of 20,000 and 50,000 steps, respectively. The learning rates for both models are set to 5e-5. During inference time, the Chart-To-Table model uses beam search with a beam width of 4.

\begin{table*}[t]
    \small
    \centering
    \begin{adjustbox}{max width=0.98\textwidth}
    {
    \begin{tabular}{lccc}
        \toprule
        
        \textbf{Dataset Split $\rightarrow~$}        & \textbf{\datashort~-\textsc{Lvlm}} & \textbf{\datashort~-\textsc{Llm}} & \textbf{\datashort~-\textsc{Ft}} \\

        \textbf{Evaluation Metric $\rightarrow~$} & GPT-4V & GPT-4V & GPT-4V  \\
        \textbf{Correction Model $\downarrow~$}\\
        \midrule
        N/A                   & \underline{50.89} & 23.47 & 24.83 \\
        LLaVA                 & 29.87 & 22.45 & 39.45 \\
        Bard                  & 37.37 & 31.77 & 44.86 \\
        GPT-4V                & \textbf{61.34} & \textbf{52.35} & \textbf{74.79}\\
        DePlot + GPT-4        & 23.79 & 22.45 & 40.63 \\
        \midrule
        \modelshort~ (Ours)   & 35.96 & \underline{39.29} & \underline{55.56} \\ 
        
        \bottomrule
    \end{tabular}
    }
    \end{adjustbox}
    \caption{Correction performance on \datashort~ using GPT-4V as the evaluation metric. GPT-4V, when used as an evaluator, assigns significantly higher scores to its own generations. This suggests potential self-enhancement bias of GPT-4V. Note that GPT-4V also assign a high scores to the original captions (i.e. N/A) on the \textsc{Lvlm} split. This is because half of these captions are directly generated from GPT-4V.} 
    \label{tab:gpt_4v_eval}
\end{table*}

\begin{figure*}[t]
    \centering
    \includegraphics[width=0.95\linewidth]{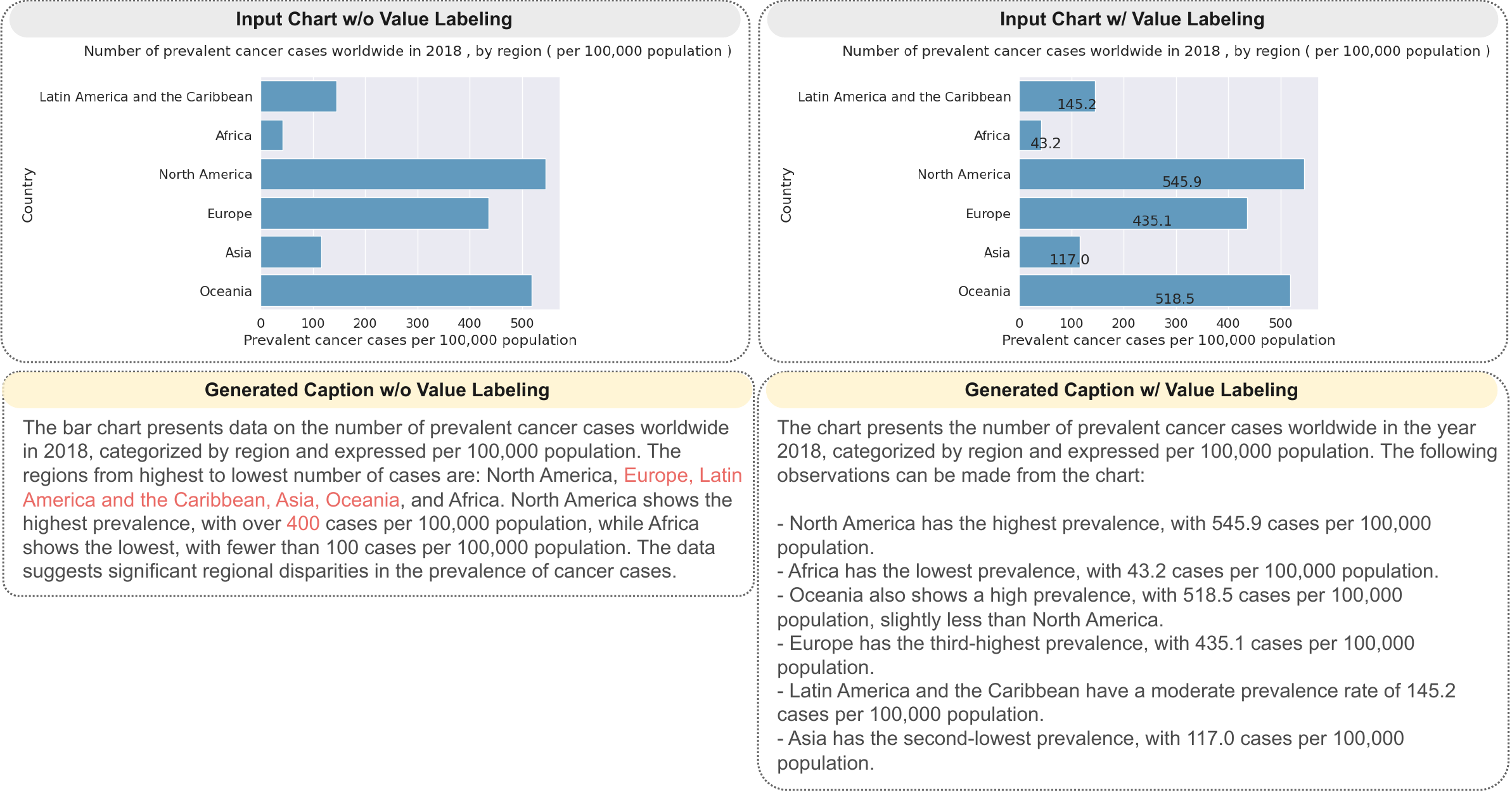}
    \vspace{-2mm}
    \caption{The impact of value labeling. We prompted GPT-4V to generate captions of two charts we created using the Seaborn library from an underlying table sampled from the Chart-to-Text dataset, with or without labeling the values of the bars on the chart. We see that when the labeled values are presented in the chart, GPT-4V is capable of producing more factual captions.}
    \vspace{-5mm}
    \label{fig:value_labeling_impact}
\end{figure*}

\begin{figure*}[b]
    \centering
    \includegraphics[width=0.95\linewidth]{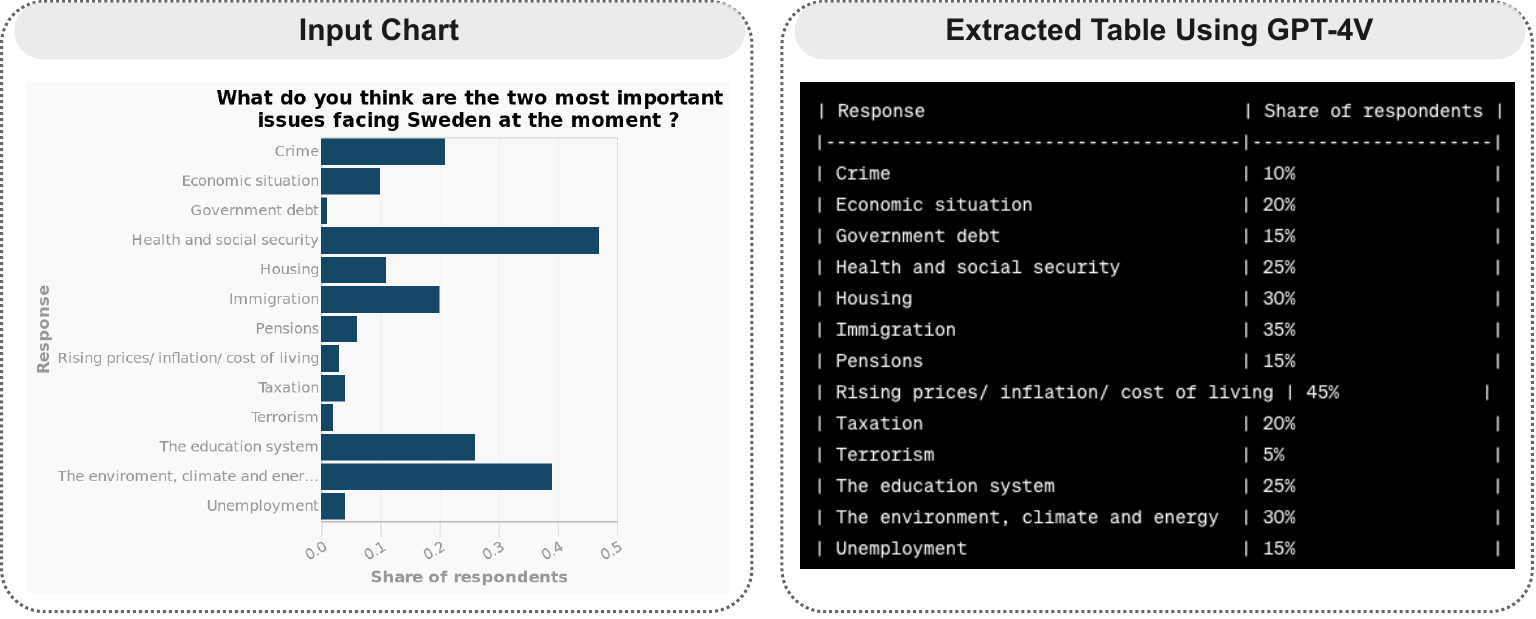}
    \caption{An example showing GPT-4V cannot accurately extract tables from charts. This indicates its inability to infer the actual value of each data point within the chart.}
    \label{fig:gpt4v_extracted_table}
\end{figure*}

\section{Prompts}
\label{apx:prompts}
\begin{figure*}[t]
    \centering
    \includegraphics[width=0.85\linewidth]{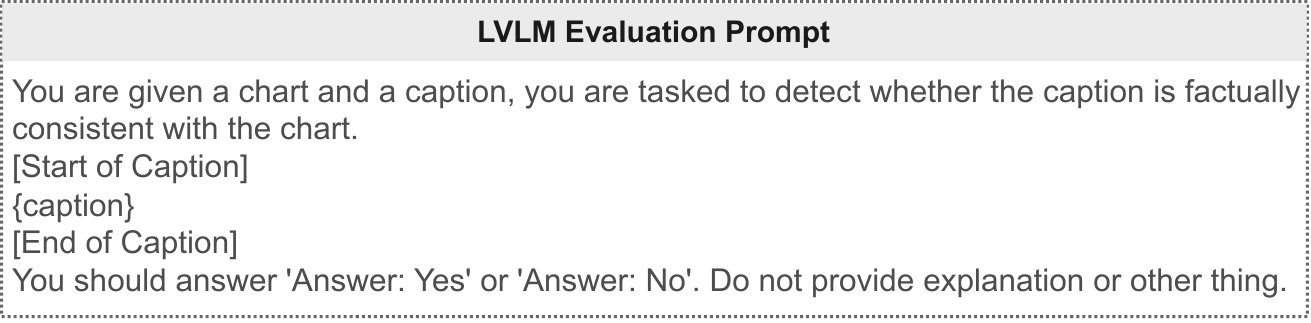}
    
    \caption{Prompts for using GPT-4V, Bard, and LLaVA-1.5 as a evaluator.} 
    
    \label{fig:lvlm_eval_prompt}
\end{figure*}

\begin{figure*}[t]
    \centering
    \includegraphics[width=0.85\linewidth]{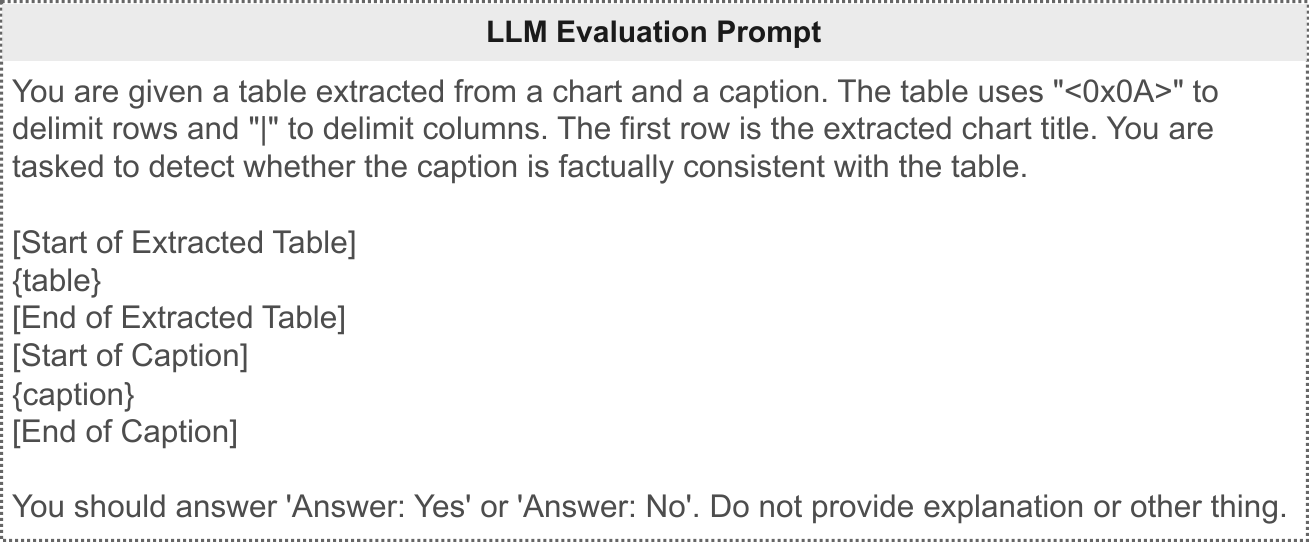}
    
    \caption{Prompts for using DePlot + GPT-4 as a evaluator.} 
    
    \label{fig:llm_eval_prompt}
\end{figure*}

\begin{figure*}[t]
    \centering
    \includegraphics[width=0.85\linewidth]{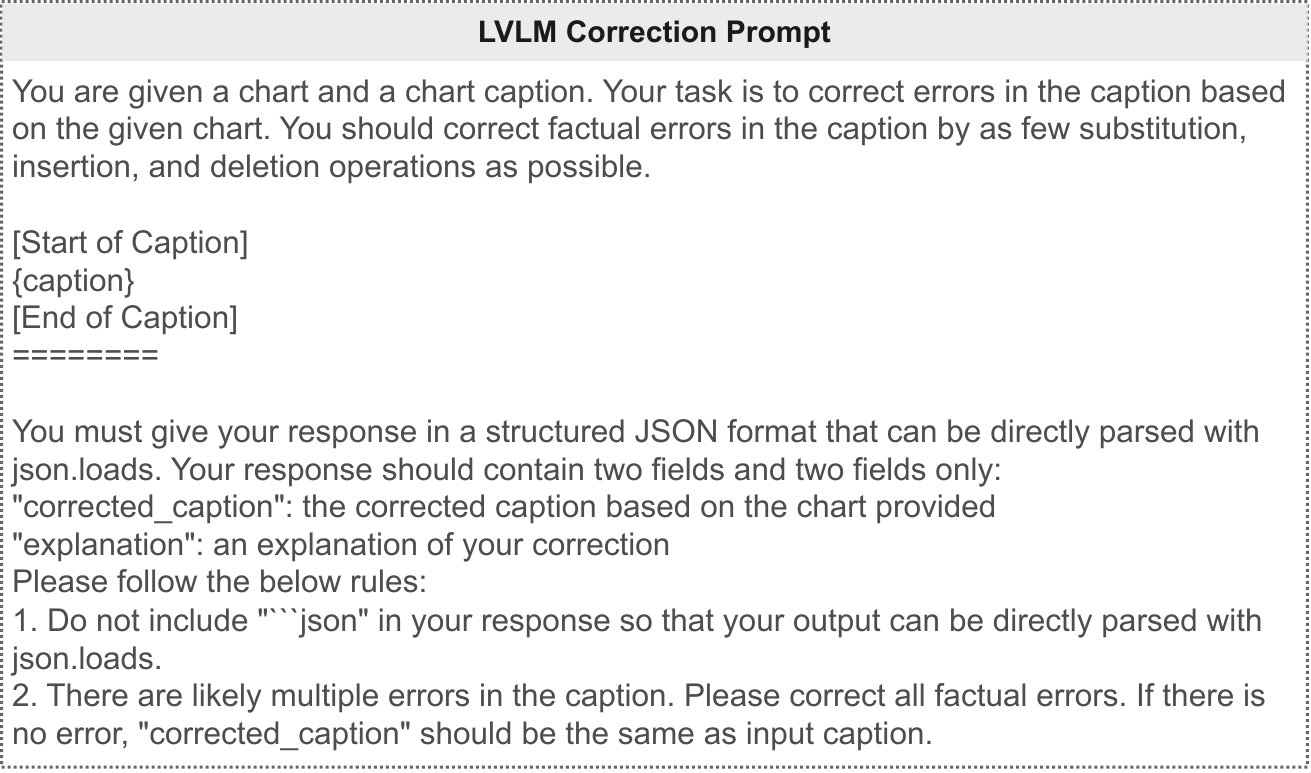}
    
    \caption{Prompts for using GPT-4V, Bard, and LLaVA as a factual error corrector.} 
    
    \label{fig:lvlm_correction_prompt}
\end{figure*}

\begin{figure*}[t]
    \centering
    \includegraphics[width=0.85\linewidth]{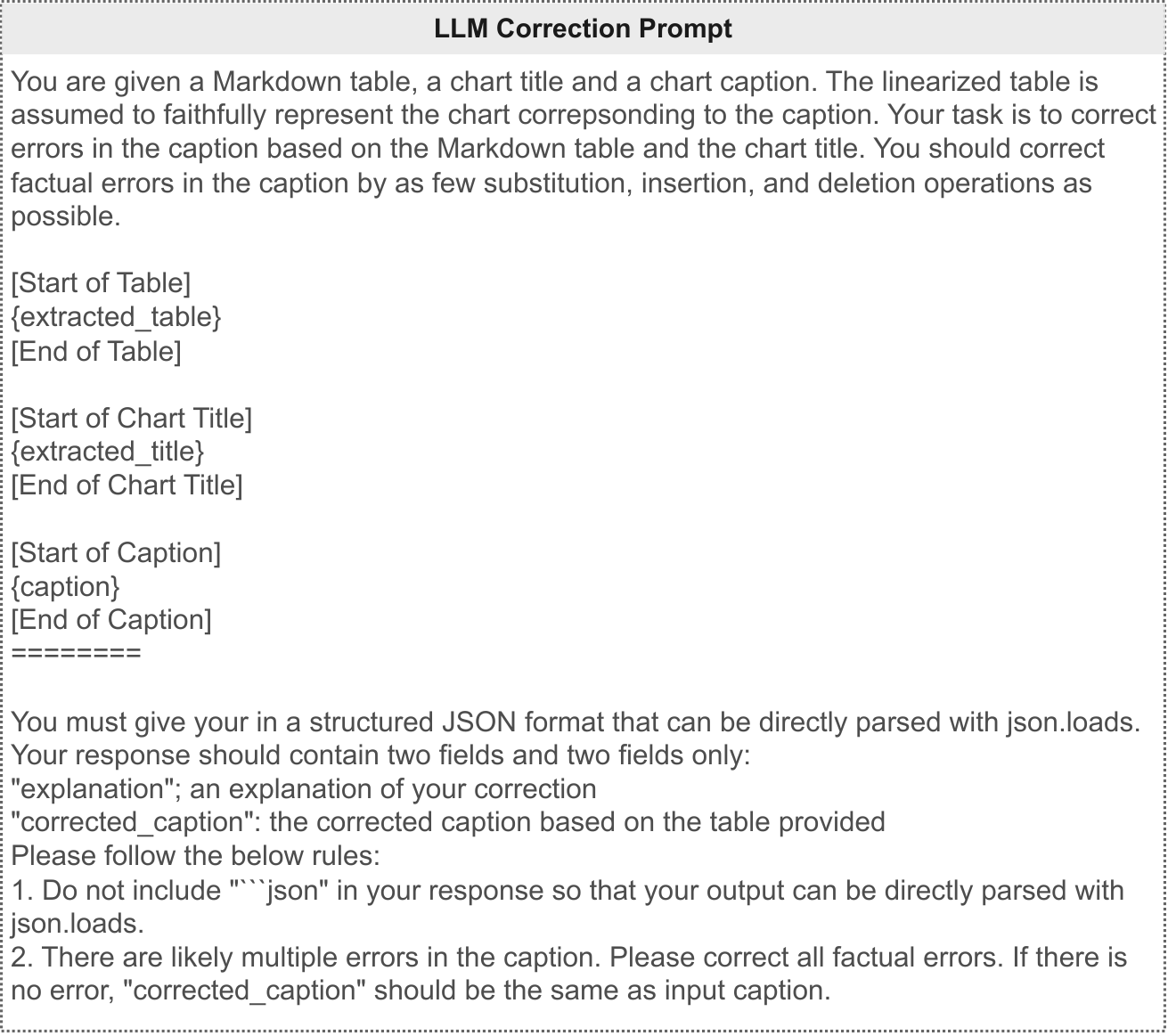}
    
    \caption{Prompts used for using DePlot + GPT-4 as a factual error corrector.} 
    
    \label{fig:llm_correction_prompt}
\end{figure*}

The prompts for using LVLM and LLM as evaluation metrics are displayed in \Cref{fig:lvlm_eval_prompt} and \Cref{fig:llm_eval_prompt}, while the prompts for factual error correction are shown in \Cref{fig:lvlm_correction_prompt} and \Cref{fig:llm_correction_prompt}.

\end{document}